\documentclass[twocolumn]{article}
\usepackage{amsmath}
\usepackage{amssymb}
\usepackage[numbers]{natbib}
\usepackage{url}
\usepackage{authblk}
\usepackage{hyperref}
\bibpunct{[}{]}{;}{a}{}{,}

\begin{document}
\title{Distance Metric Ensemble Learning\\and the Andrews-Curtis Conjecture}
\author[1]{Krzysztof Krawiec\thanks{krawiec@cs.put.poznan.pl}}
\author[2]{Jerry Swan\thanks{jerry.swan@york.ac.uk}}
\affil[1]{Poznan University of Technology, Piotrowo 2, 60-965 Poznan, Poland.}
\affil[2]{University of York, Deramore Lane, York, YO10 5GH, UK.}

\maketitle

\begin{abstract}
Motivated by the search for a counterexample to the Poincar\'e conjecture in three and four dimensions, the Andrews-Curtis conjecture was proposed in 1965. It is now generally suspected that the Andrews-Curtis conjecture is false, but small potential counterexamples are not so numerous, and previous work has attempted to eliminate some via combinatorial search. Progress has however been limited, with the most successful approach (breadth-first-search using secondary storage) being neither scalable nor heuristically-informed. A previous empirical analysis of problem structure examined several heuristic measures of search progress and determined that none of them provided any useful guidance for search. In this article, we induce new quality measures directly from the problem structure and combine them to produce a more effective search driver via ensemble machine learning. By this means, we eliminate 19 potential counterexamples, the status of which had been unknown for some years.
\end{abstract}

\section{Introduction}

The Andrews-Curtis conjecture (ACC) \citep{AndrewsCurtis:1965} dates
back to 1965 and is an open problem of widespread interest in low-dimensional
topology \citep{Wright, Hog-AngeloniMetzler} and combinatorial group
theory \citep{BurnsMacedonska1993,MillerSchupp:1999}. It originated
in the search for a counterexample to the Poincar\'e conjecture in three
and four dimensions. Subsequent to the proof of the Poincar\'e conjecture
\citep{citeulike:10386641}, it is generally suspected that ACC is
false. Attention has therefore shifted to potential counterexamples
to ACC, of which relatively few of likely computational tractability are known \citep{Bridson}.

ACC can be stated in both group-theoretic and topological terms. We
proceed via the elementary theory of group presentations \citep{Johnson90}.
%A group is a set of \emph{generators} on which is defined a closed binary operation `{*}' that is invertible, associative and has an identity element. A \emph{word} is a sequence of generators, with juxtaposition and exponentiation used as shorthand for composition, i.e.\ for generators $a,b$, the word $ab^{2}a$ denotes $a*b*b*a$. A word is \emph{freely reduced} if no generator is adjacent to it's inverse and \emph{cyclicly reduced} if it is not of the form $g^{\pm1}wg^{\mp1}$ for any generator $g$ or subword $w$. A (finite) \emph{presentation} of a group is defined by additionally specifying a set of \emph{relators}, i.e.\ a set of words that are known to be equivalent to the identity element of the group, the latter being represented by the empty word. For example, the relator $a^{3}$ expresses that the sequence of generators $a*a*a$ is equal to the empty word. Presentations are denoted in the form $\langle\textrm{{\it {generators}}}|\textrm{{\it {relators}}}\rangle$,e.g.\ $\langle a|a^{3}\rangle$ denotes the cyclic group of order 3. 
A finite presentation $\langle g_{1},\ldots,g_{m}|r_{1},\ldots,r_{n}\rangle$
is said to be \emph{balanced} if $m$ is equal to $n$. The \emph{trivial
presentation of the trivial group} of \emph{rank} $r$ is the balanced
presentation $\langle g_{1},\ldots,g_{r}|g_{1},\ldots,g_{r}\rangle$. For conciseness, we sometimes denote the inverse of a generator by capitalization, e.g.\ $B=b^{-1}$. 

The group-theoretic version of ACC states that ``every balanced presentation
of the trivial group can be transformed into the trivial presentation
via some sequence of \emph{AC-moves}'' \citep{BurnsMacedonska1993}.
For relators $r_{i},r_{j}$, the {AC}-moves are: 
\begin{enumerate}
\item \textbf{AC1.} \ $r_{i}\rightarrow{r_{i}}^{-1}$ (inversion of a relator) 
\item \textbf{AC2.} \ $r_{i}\rightarrow r_{i}r_{j},i\neq j$ (multiplication
of one relator by another) 
\item \textbf{AC3.} \ $r_{i}\rightarrow g^{\mp1}r_{i}g^{\pm1}$, (\emph{conjugation}
of a relator by some generator $g$) 
\end{enumerate}
We say that a sequence of AC-moves that connects a source presentation
$p$ to the trivial group is an \emph{AC-trivialisation} of $p$. The contribution of this article is a novel approach to searching for AC-trivializations, leading to the elimination of 19 of the potential counterexamples described at the end of the next section. The proposed metaheuristic algorithm combines offline learning and online ensemble approaches \cite{667881}.

%In implementation terms, the concrete representation of the above is considerably more simple than the above definition might suggest to those unfamiliar with group theory: a balanced presentation can be represented as a list of words, with words being conveniently represented as lists of integers excluding zero (representing the generators) with the property that $g^{-1}$ is represented by $-g$.
% Moved from the end of Section 2 and modified a bit. 
% JS: All good. DONE.

\section{Previous Work\label{sec:Previous-Work}}

It is possible to investigate potential counterexamples to ACC using
combinatorial search techniques such as genetic algorithms \citep{Holland:1992:ANA:129194}
and breadth-first search \citep{HavasRamsay:2003,BowmanMcCaul:2006}.
The search is therefore for some sequence of moves connecting the
trivial group to a potential counterexample. Metaheuristic approaches are guided by a \emph{fitness function}, an ordering % Unless you have sth special in mind, most fitness functions (including those used for ACC AFAIR) are scalar, and thus induce *complete* orders. Did you mean *preorder* here?
% JS: For all practical purposes you are correct - I've historically used a partial ordering so `unfeasible' solutions can be unambiguously distinguished by some sentinel value. DONE.
on solution states that gives a heuristic measure
of the quality of a solution. Despite both group-theoretic \citep{MiasnikovMyasnikov:2003}
and metaheuristic \citep{Miasnikov:1999} approaches, the state-of-the-art
since 2003 has been breadth-first search \citep{HavasRamsay:2003},
subsequently extended to efficiently index secondary storage \citep{BowmanMcCaul:2006}.
More recently, \citep{DBLP:journals/tinytocs/Lisitsa13}
used the alernative approach of first-order theorem proving to obtain
trivializations for all previously-eliminated potential counterexamples.

The AC-moves themselves form a group (denoted ${AC}_{n}$) under their
action on balanced presentations of rank $n$: AC1 and AC3 are self-inverse
and AC2 is inverted by multiplication by the inverse of the source
relator. One can therefore either start from a potential counterexample
and search `forwards' towards the trivial presentation or else start
at the trivial presentation and apply inverse moves. In this manner,
Havas and Ramsay make use of bidirectional breadth-first search \citep{HavasRamsay:2003},
terminating with success if the search frontiers intersect. For a
balanced presentation of rank $n$, there are $3n^{2}$ {AC}-moves,
and hence % Not sure if the term 'upper bound' is appropriate here. To me, all those sequences are 'valid', but some of them are 'partially ineffective'. I'm rephrasing this; the previous version is saved below. 
% JS: Let's go with your version ;-). DONE.
$(3n^{2})^{l}$ move sequences of length
$l$. By the group property of $AC_{n}$, it is clear that the effective length of many sequences is lower, e.g.\ the immediate re-application
of self-inverse moves will always yield the previously-encountered
presentation. In practice, for the rank 2 case, Havas and Ramsay note
that the theoretical branching factor of 12 tends to average at
around 8 in the low-depth unconstrained investigations they performed.
\iffalse % previous version
and hence an upper bound of $(3n^{2})^{l}$ move sequences of length
$l$. By the group property of $AC_{n}$, it is clear that such an
upper bound will not be achieved: e.g.\ the immediate re-application
of self-inverse moves will always yield the previously-encountered
presentation. In practice, for the rank 2 case, Havas and Ramsay note
that the theoretical branching factor of 12 tends to an average at
around 8 in the low-depth unconstrained investigations they performed.
\fi

For breadth-first search, constraints on relator length are used to
make the state space finite (and furthermore tractable), and `total
length of relators' has been used as an estimate of problem difficulty.
In these terms, the smallest potential counterexample is ${AK}_{3}=\langle a,b|a^{3}B^{4},abaBAB\rangle$
of length 13 due to Akbulut and Kirby \citep{AkbulutKirby:1985}.
In \citep{BowmanMcCaul:2006} Bowman and McCaul exhaustively enumerated
the constrained search space for $AK_{3}$ for maximum individual
relator lengths from 10 to 17 inclusive, but were unable to find a
solution sequence, despite enumerating 85 million presentations and
taking 93 hours on an IBM z800 mainframe. %It is part of the folklore of the Andrews-Curtis conjecture that the worst case length of trivialization sequences is a superexponential function of relator length \citep{BridsonACSuperexponential},
% JS: slight rewording of the next sentence:
It is clearly therefore necessary to explore alternative approaches. In this paper, we investigate the application of a more informed version of metaheuristic search than has previously been attempted. Metaheuristic search has an associated \emph{fitness landscape} \citep{Wright:32,Stadler:1995},
i.e.\ a graph in which the vertices are (potential) solutions and the
edges of the graph represent the operations for transforming a solution
into its neighbour. Previous work by Swan et al.\ \citep{doi:10.1142/S0218196711006753}
has explored alternatives to relator length as a fitness measure (e.g.\ edit distance) and determined that fitness does not correlate well with the distance (expressed in terms of number of edges traversed) to a solution \citep{JonesForrest:1995}. 

%KK: moved here from Conclusions
In a broader context, with the exception of work by Spector et al.\ \citep{Spector:2008:gecco} which uses genetic programming \citep{Koza:1992:GPP:138936} to discover terms with specific properties in finite universal algebras, we are not aware of any significant applications of machine learning techniques to algebraic problems of general interest. 

\section{Problem Instances}
Determining if a balanced presentation actually represents the trivial
group (and is therefore a potential counterexample) is a nontrivial
task \citep{EdjvetEtAl2001,MiasnikovMyasnikov:2003}
in its own right. A regularly-updated collection of balanced presentations
arising from computational and algebraic investigations into irreducible cyclicly presented groups performed since 2001
\citep{Edjvet2003,Edjvet2011165,doi:10.1142/S0218196710005777,icpl17}
is maintained at \citep{EdjvetHomepage}. These fall into two categories: 
\begin{itemize}
\item Presentations $T_{i}$ known to be trivial. These are potential counterexamples
to ACC, and are of interest as described above. 
\item Presentations $O_{i}$ for which triviality is an open question. In
such cases, obtaining an AC-trivialisation additionally provides answers to a
question of longstanding interest due to Dunwoody \citep{Dunwoody1995}. 
\end{itemize}
These instances have resisted further investigation by both algebraic
and computational approaches over a number of years. For the instances
$O_{i}$, approaches have included application of string-rewriting systems via the automatic groups software packages \textsc{KBMAG} \citep{Holt1995} and \textsc{MAF}
\citep{Williams2010} and the computer algebra package \textsc{Magma}
(via the algorithms for simple quotients or low-index subgroups \citep{BosmaEtAl1997}).

%DONE: I prefer to have it at the end of Intro. I'm afraid many readers will miss this 'mission statement' hidden at the end of review section. Also, it was not clear what kind of 'functions' you mean here.
%The contribution of this article is a novel approach to generating such functions that combines offline learning and online ensemble approaches. We use this technique to eliminate some of the potential counterexamples described above.
% JS: all good by me. DONE.
\section{Methodology\label{sec:Methodology}}

As discussed above, in combinatorial terms, algorithmic verification of AC counterexamples
is a search problem, with states corresponding to presentations
and neighborhood defined by the set of AC moves. No fitness
function is known that would efficiently guide search in this space,
and, as shown in \citep{doi:10.1142/S0218196711006753}, the functions
used in past studies do not correlate well with the actual distance
to the search target (i.e.\, the number of AC moves required to reach
the trivial presentation). It is therefore unsurprising that the greatest
successes to date have not been heuristically informed. One of the conclusions of Swan et al. was that an adaptive or penalty-driven fitness function may allow metaheuristic approaches to outperform breadth-first search.

The central claim of this study is that, in the likely absence of
a unique, global and efficiently-computable fitness % DONE:  \footnote{We should either stick with `fitness' or constrast fitness and objective functions}
% KK: Right. You've elegantly introduced fitness earlier, so we will stick to it. I've changed to 'fitness' everywhere - except when we talk about single- and multiobjective variants, where obviously 'multi-fitness' would sound weird. DONE.
 function, a potentially
useful substitute for it can be \emph{learned} from the problem. In
the following, we refer to such substitute functions as \emph{distance
metrics}. A distance metric should exhibit (some degree of) correlation
with the actual (in practice unknown) distance from the search target.
In contrast to conventional fitness functions, we do not require
it to be globally minimized at the search target. In outline, the method is split into three phases: 
\begin{enumerate}
\item Preparation of a training set of presentations.
\item Offline learning of a set of distance metrics on presentations. 
\item Online search for trivialising sequences of AC-moves, using genetic
algorithms equipped with a fitness measure which is informed by an
ensemble of the generated distance metrics. 
\end{enumerate}
Of these phases, detailed in subsequent sections, phases 2 and 3 are
implemented as a generational genetic algorithm \citep{Holland:1992:ANA:129194}. 

\subsection{Preparation of training set}\label{sec:prepCases}

This phase consists in generation of a training set of \emph{fitness
cases}, i.e., examples with which the distance metrics are trained.
Each fitness case is a pair $(p,l)$, where $p$ is a randomly-chosen
presentation a small number of AC-moves from the trivial presentation,
and $l$ is the length of the shortest path that trivialises $p$
(referred to as \emph{distance} in the following). It is clearly not
possible in general to obtain this path directly by starting at some
arbitrary $p$, since this would be equivalent to showing that $p$
is AC-trivializable. This forces us to devise a different approach for drawing the presentations
for fitness cases. We start from the trivial presentation $t$, perform
a reverse random walk, and terminate it at a state $p$ if walk length exceeds $60$ or the total length of relators reaches $60$. 
% given constraints\footnote{ are these always relator length? If so, let's just say so here.}) KK: DONE. From what I found in the parameter files, walk length was constrained to [6,60], and total relator length to 60 
% \footnote{KK: --- Can you explicitly add relator constraints to the `parameters' section?} KK: DONE, but this is probably the best location to mention them. 
Then, we attempt to find the corresponding inverse
forward walk from $p$ to $t$.  Since all AC-moves other than multiplication
are self-inverse, the length of forward and reverse walks will generally
correlate well, however a move such as $(AABaB,BAbaB)\rightarrow(AB,BAbaB)$
from the proof of the ${AK}_{2}$ example from \citep{HavasRamsay:2003}
requires several moves to invert.

In principle, we could perform the reverse random walk by simply applying
a random sequence of AC-moves of a given length to the trivial presentation.
Since such a walk is unlikely to be the shortest path, we instead
explicitly build the graph of reverse moves rooted at the trivial
presentation using breadth first search and subsequently sample walks
from it. 

The outcome of this stage is a set of fitness cases $T=\{(p,l)\}$.
For all presentations of a given rank, it is sufficient to produce
such a sample once, as all instances of potential ACC counterexamples
dwell in the same search space. 

\subsection{Offline Distance Learning\label{sub:distLearning}}

Given a sample of fitness cases $T$, the goal of the next step is
to learn an approximate unary distance metric $d:P\rightarrow\mathbb{R}^{+}$
that, for a given presentation $p$, predicts its distance from the
trivial presentation $t$. Ideally, we would like to synthesize a
function $d$ such that $d(p)=l$ for every fitness case $(p,l)$
in $T$ (and possibly beyond it), but this cannot be done otherwise
than by running a costly tree search from $p$ and terminating once
$t$ has been reached. Rather than that, we attempt to learn \emph{heuristic
estimates} of $l$. Based on this motivation, we set our goal to
learning $d$ such that $d(p)$ and $l$ are well \emph{correlated}. 

Technically, the process of learning the distance metric is realized as
an evolutionary algorithm working with the population of \emph{candidate
solutions,} each of them representing a specific distance measure
$d$. The fitness value of a given candidate solution $d$ is calculated
by applying $d$ to all fitness cases in $T$ and computing the correlation
coefficient between $d(p)$ and $l$. Depending on the setup, we employ
linear correlation (Pearson) or rank-wise correlation (Kendall). 

The heuristic distance estimates learned in this way are universal
in representing domain knowledge that is common for all ACC instances
of a given rank. This is another reason for treating this learning
process as a separate stage (stage 2) of our workflow that precedes
the actual solving of particular instances of ACC (Section \ref{sub:Online-search}). For the same reason, we refer to it as to \emph{offline} distance
learning. 

The critical design choice concerns the representation of distance
metrics. Previous studies resorted to total sum of relator lengths
of $p$, edit distance between $p$ and the trivial presentation $t$,
or other generic metrics (see Section \ref{sec:Previous-Work}). Following \citep{doi:10.1142/S0218196711006753},
we posit that no significant correlation with the actual distance
can be achieved without involving a greater degree of domain knowledge. On
the other hand, manual design of metrics is time consuming, and likely
to result in measures which suffer from unhelpful bias. 

We therefore elected to represent the candidate metrics in the same manner as
the solutions to the underlying ACC problem: each $d$ is a sequence
of ACC moves. When evaluated on a given presentation $p$, the moves
in $d$ are applied to $p$ one by one, resulting in a certain presentation
$p'$. The total relator length of the resulting presentation $|p'|$
is interpreted as the value of $d(p)$. The correlation of $d(p)$
with $l$ forms the fitness of the candidate metrics. 

The objective of this evolutionary distance learning is thus to synthesize
a sequence of moves that `corrects' the total relator length of a
given presentation w.r.t.\ its actual distance from $t$ (i.e., appropriately
shortens and extends the relators). By resorting to correlation, we
do not require the total relator length of the resulting presentation
$p'$ to be \emph{equal} to the actual distance. 

However, even with this relaxation, it would be na\"ive to assume that
a metric with perfect correlation (over the set of all starting presentations
$p$ available in $T$) can be expressed using AC moves. Thus, rather
than searching for a single ideal metric, we run evolutionary search
$50$ times and collect the best-of-run candidate metrics from
all runs, forming so a \emph{sample of metrics} $D$.
The distance metrics obtained in particular runs are gathered in a
set $D$ that is a parameter of the subsequent step. This is an example of ensemble learning \citep{667881}, in which the deficiencies of inaccurate predictors are improved by generating a diverse collection of them and aggregating their outputs. The resulting ensemble is then expected to have greater accuracy than an individual predictor. 

\subsection{Online search\\for AC-trivialisations\label{sub:Online-search}}

The set of distance metrics $D$ learned in the previous section allows us to devise fitness functions to guide the actual search for trivializations. In contrast to the previous two steps, this stage proceeds \emph{online}, i.e., for each presentation (problem instance) independently.

Metaheuristic search is parameterized by three essential components,
viz.\ a solution representation, a set of operators for changing
or recombining solution representations, and a a fitness measure.
We adopt the same formulation for the first two of these as the genetic
algorithms approach of Miasnikov \citep{Miasnikov:1999}, i.e.\ solutions
are represented as sequences of AC-moves and operators are the insertion,
deletion and substitution of a move. These operators take one solution as an argument and are by this token known as \emph{mutations} in the terminology of genetic algorithms. No binary (two-argument) \emph{crossover} search operators are applied in our setup. %\footnote{JS: DONE Do we use crossover? If so, we should give crossover probability somewhere, or else state that it isn't used}. KK: No we don't. We might have tried it in the very beginning of our experimentation. I've added a remark on the lack of crossover here, but maybe it's too verbose.  

Our fitness measure is based on the approximate metrics
learned in the process described in Section \ref{sub:distLearning}.
We consider two ways of conducting the selection process based on the set % was: defining a fitness function based on the set. JS: DONE.
of metrics $D$ learned there. \medskip

\textbf{Single objective}. In this variant, we apply the metrics in
$D$ to the training sample $T$ of fitness cases and perform multiple
linear regression of the obtained values against the reverse walk length
 $l$. In other words, a vector $\mathbf{w}$ of weights
$w_{i}$ is found such that the linear combination of the distances
\begin{equation}
f(p)=\sum_{d_{i}\in D}w_{i}d_{i}(p)\label{eq:scalarFitness}
\end{equation}
minimizes the square root error with respect to $l$, i.e.\, 
\[
\min_{\mathbf{w}}\sum_{(p,l)\in T}(f(p)-l)^{2}.
\]
The function $f(p)$ constructed in this way becomes the fitness that
drives a conventional single-objective search as in the approach of Miasnikov
\citep{Miasnikov:1999}. 
%As is common in genetic algorithm configuration, in the selection stage of an evolutionary run, candidate solutions $f(p)$ are chosen by competing via tournament selection (of size $7$). \medskip{}
% Chris: I've commented this out, since we mention it below. KK: Good. DONE. 

\textbf{Multi-objective}. Recent work in evolutionary computation
indicates that heuristic search can be more effective when driven
with multiple objectives (fitness functions) rather than one \citep{DBLP:journals/jmma/Jensen04}. 
%\footnote{any good ref for this?}. DONE
Simultaneously maximizing  multiple objectives that express various characteristics of candidate solutions is a natural means for maintaining population diversity and reduces the risk of premature convergence, i.e.\, all candidate solutions in the population becoming very similar to each other (which hinders explorations of the search space). 

Following these observations, in the second variant we do not combine
the particular metrics $d_{i}\in D$ into a common fitness as in
(\ref{eq:scalarFitness}), but treat every $d_{i}\in D$ as a separate
objective. In the selection stage of evolutionary run, we use Non-dominated
Sorting Genetic Algorithm (NSGA-II, \citep{Deb2002}). Given a population, NSGA-II builds a Pareto-ranking based on dominance relation that spans the objectives, and then employs tournament selection on Pareto-ranks to select the solutions. Given two solutions with the same Pareto-rank, it prefers the one from the less `crowded' part of Pareto-front.  
%\footnote{Chris - this sentence is a bit dense, can you maybe rephrase or unpack it a bit?}, DONE: Is this better? JS: All good. DONE.

It is a known that multiobjective selection methods like NSGA-II tend to become ineffective when the number of objectives is high. Given the $50$ objectives gathered in $D$, it is very unlikely for any candidate solution (move sequence) to dominate on these objectives any other move sequence in a working population. In order to reduce the number of objectives used in the multiobjective variant, we employ a heuristic procedure that trims $D$ to the $5$ least correlated objectives, % search drivers\footnote{Chris - to be consistent, shall we instead say `fitness functions' here?}, DONE. 
where correlation is calculated in the same manner as in Section \ref{sub:distLearning}, i.e.\ with respect to the sample of presentations prepared in Section \ref{sec:prepCases}.  \medskip{} 

In both single- and multi-objective scenarios, we employ settings which are quite conventional for evolutionary algorithms. The initial population of size $1000$ is seeded with random sequences of length $8$. In each iteration (generation), tournament selection with tournament of size $7$ is applied to appoint the `parent' candidate solutions that are then modified by search operators. In the single-objective variant, the selection is based on the scalar objective, while in the multi-objective variant it works with the ranks in the Pareto ranking induced by the dominance relation. 

The selected solutions undergo one of three possible modifications (\emph{search operators}), with the accompanying probabilities: 
\begin{itemize}
\item Insertion of a randomly selected AC-move at a random location of a sequence (prob. 0.1).
\item Replacement of a move at random location with a randomly generated AC-move (prob. 0.8). 
\item Deletion of a move at random location (prob. 0.1). 
\end{itemize}
Thanks to equal probability of insertion and deletion, the expected change of length of this suite of operators is zero. Nevertheless, preliminary experiments showed that longer sequences tend to obtain better fitness. Therefore, to prevent excessive growth, sequences longer than $70$ moves are assigned the worst possible fitness (which almost always results in eliminating them from a population). On the other hand, to avoid wasting time on considering very short sequences that are unlikely to trivialize the presentation in question, we penalize sequences shorter than $8$ moves in the same way. Finally, the same penalization is applied to the sequences that traverse presentations with the total relator length greater or equal to $200$.  

Search proceeded until a trivializing sequence was found, or the number of generations reached $100,000$, or three-hour runtime elapsed, whatever came first. As the evolutionary search is stochastic by nature and depends on the choice of initial population, we repeated the runs for every presentation for 
%at least - JS: DONE. would it not be better to fix on a single specific value? KK: Maybe I was not clear enough here. We use the same vector of 20 seeds for every presentation. I've removed the 'different' adjective here, if that helps. 
% JS:All good. DONE.
$20$ seeds of a random number generator. The entire experiment involved at least $10,000$ evolutionary runs in total. The computations were conducted on a cluster of workstations equipped with $4$-core CPUs, running under the Simple Linux Utility for Resource Management (SLURM) software framework \citep{Jette02slurm:simple}. 

With regard to the balanced presentations upon which AC move sequences
act, we adopt two canonicalization constraints that differ slightly from those
used in \citep{BowmanMcCaul:2006}, defined as follows: 
\begin{itemize}
\item \textbf{C1}. Relators are sorted in shortlex, i.e.\ `length then lexicographic' order. 
\item \textbf{C2}. Relators are chosen to be the least representative (under
shortlex ordering) modulo cyclic permutation and inversion, subject to the constraint that it is freely reduced. This weaker constraint is necessary since we cannot enfore cyclic reduction: to do so would obviate the AC3 conjugate moves. 
\end{itemize}
These constraints reduce the size of the search space in the graph-
and walk- generation phases, albeit at additional computational expense
in sorting and determining equivalence.

\section{Results}

We conduced an extensive series of computational experiments on the
$T_{i}$ and $O_{i}$ presentations introduced at the end of Section
\ref{sec:Previous-Work}, using several variants of the workflow presented
in Section \ref{sec:Methodology}. Table \ref{tab:solvedPresentations}
presents the list of presentations that have been solved by this setting,
i.e., demonstrated to be AC-trivializable. 

\begin{table*}
\protect\caption{List of presentations solved (AC-trivialised) by the proposed approach.\label{tab:solvedPresentations}}
\center\begin{tabular}{|c|c|c|}
\hline 
Identifier & Presentation & Trivialization Length\\
\hline 
\hline 
T1 & $\langle x_{0},x_{1}|x_{0}^{2}x_{1}X_{0}X_{1},x_{1}^{2}x_{0}X_{1}X_{0}\rangle$ & 6\\
\hline 
T5 & $\langle x_{0},x_{1}|x_{0}^{2}x_{1}^{2}X_{0}X_{1}^{2},x_{1}^{2}x_{0}^{2}X_{1}X_{0}^{2}\rangle$ & 10\\
\hline 
T11 & $\langle x_{0},x_{1}|x_{0}^{3}x_{1}^{2}X_{0}^{2}X_{1}^{2},x_{1}^{3}x_{0}^{2}X_{1}^{2}X_{0}^{2}\rangle$ & 14\\
\hline 
T13 & $\langle x_{0},x_{1}|x_{0}^{2}x_{1}X_{0}x_{1}X_{0}X_{1},x_{1}^{2}x_{0}X_{1}x_{0}X_{1}X_{0}\rangle$ & 7\\
\hline 
T29 & $\langle x_{0},x_{1}|x_{0}^{3}x_{1}^{3}X_{0}^{2}X_{1}^{3},x_{1}^{3}x_{0}^{3}X_{1}^{2}X_{0}^{3}\rangle$ & 21\\
\hline 
T31 & $\langle x_{0},x_{1}|x_{0}^{3}x_{1}X_{0}x_{1}X_{0}X_{1}^{2},x_{1}^{3}x_{0}X_{1}x_{0}X_{1}X_{0}^{2}\rangle$ & 10\\
\hline 
T34 & $\langle x_{0},x_{1}|x_{0}^{2}x_{1}^{2}x_{0}X_{1}X_{0}^{2}X_{1},x_{1}^{2}x_{0}^{2}x_{1}X_{0}X_{1}^{2}X_{0}\rangle$ & 10\\
\hline 
T35 & $\langle x_{0},x_{1}|x_{0}^{2}x_{1}^{2}X_{0}x_{1}X_{0}X_{1}^{2},x_{1}^{2}x_{0}^{2}X_{1}x_{0}X_{1}X_{0}^{2}\rangle$ & 24\\
\hline 
T39 & $\langle x_{0},x_{1}|x_{0}^{2}x_{1}X_{0}x_{1}^{2}X_{0}X_{1}^{2},x_{1}^{2}x_{0}X_{1}x_{0}^{2}X_{1}X_{0}^{2}\rangle$ & 10\\
\hline 
T56 & $\langle x_{0},x_{1}|x_{0}^{4}x_{1}^{3}X_{0}^{3}X_{1}^{3},x_{1}^{4}x_{0}^{3}X_{1}^{3}X_{0}^{3}\rangle$ & 25\\
\hline 
T61 & $\langle x_{0},x_{1}|x_{0}^{3}x_{1}^{2}X_{0}x_{1}X_{0}^{2}X_{1}^{2},x_{1}^{3}x_{0}^{2}X_{1}x_{0}X_{1}^{2}X_{0}^{2}\rangle$ & 14\\
\hline 
T63 & $\langle x_{0},x_{1}|x_{0}^{3}x_{1}^{2}X_{0}X_{1}^{3}X_{0}x_{1},x_{1}^{3}x_{0}^{2}X_{1}X_{0}^{3}X_{1}x_{0}\rangle$ & 24\\
\hline 
T66 & $\langle x_{0},x_{1}|x_{0}^{3}x_{1}X_{0}^{2}x_{1}^{2}X_{0}X_{1}^{2},x_{1}^{3}x_{0}X_{1}^{2}x_{0}^{2}X_{1}X_{0}^{2}\rangle$ & 14\\
\hline 
T67 & $\langle x_{0},x_{1}|x_{0}^{3}x_{1}X_{0}x_{1}^{2}X_{0}X_{1}^{3},x_{1}^{3}x_{0}X_{1}x_{0}^{2}X_{1}X_{0}^{3}\rangle$ & 22\\
\hline 
T76 & $\langle x_{0},x_{1}|x_{0}^{2}x_{1}x_{0}x_{1}X_{0}X_{1}X_{0}X_{1},x_{1}^{2}x_{0}x_{1}x_{0}X_{1}X_{0}X_{1}X_{0}\rangle$ & 10\\
\hline 
T81 & $\langle x_{0},x_{1}|x_{0}^{2}x_{1}X_{0}x_{1}X_{0}X_{1}x_{0}X_{1},x_{1}^{2}x_{0}X_{1}x_{0}X_{1}X_{0}x_{1}X_{0}\rangle$ & 19\\
\hline 
T82 & $\langle x_{0},x_{1}|x_{0}^{2}x_{1}X_{0}X_{1}x_{0}x_{1}X_{0}X_{1},x_{1}^{2}x_{0}X_{1}X_{0}x_{1}x_{0}X_{1}X_{0}\rangle$ & 10\\
\hline 
T84 & $\langle x_{0},x_{1}|x_{0}^{2}X_{1}x_{0}x_{1}X_{0}x_{1}X_{0}X_{1},x_{1}^{2}X_{0}x_{1}x_{0}X_{1}x_{0}X_{1}X_{0}\rangle$ & 15\\
\hline 
T85 & $\langle x_{0},x_{1}|x_{0}x_{1}x_{0}x_{1}X_{0}^{2}X_{1}x_{0}X_{1},x_{1}x_{0}x_{1}x_{0}X_{1}^{2}X_{0}x_{1}X_{0}\rangle$ & 24\\
\hline 
\end{tabular}
\end{table*}

As can be seen from Table \ref{tab:solvedPresentations}, the obtained
AC-trivialization sequences vary in length from 6 to 25. Their lengths
prevent them from being presented here in full, so here we list
here only the sequences for presentations T1 and T13 in Fig. \ref{fig:trivExamples}, with the others
available online\footnote{\url{http://www.cs.put.poznan.pl/kkrawiec/wiki/?n=Site.AndrewsCurtis}}. For brevity, we relabel as follows: $x_{0}\mapsto a,X_{0}\mapsto A,x_{1}\mapsto b,X_{1}\mapsto B$.

\begin{figure*}
%\begin{figure}
\textbf{T1}:\\
\begin{align*} & \langle a,b | a^2bAB,b^2aBA \rangle \xrightarrow[(b^2aBA)^A]{} \langle a,b | a^2bAB,ab^2aBA^2 \rangle\\ & \langle a,b | a^2bAB,ab^2aBA^2 \rangle \xrightarrow[ab^2aBA^2 *= a^2bAB]{} \langle a,b | ab,a^2bAB \rangle\\ & \langle a,b | ab,a^2bAB \rangle \xrightarrow[(a^2bAB)^b]{} \langle a,b | ab,Ba^2bA \rangle\\ & \langle a,b | ab,Ba^2bA \rangle \xrightarrow[(ab)^A]{} \langle a,b | a^2bA,Ba^2bA \rangle\\ & \langle a,b | a^2bA,Ba^2bA \rangle \xrightarrow[(a^2bA)^{-1}]{} \langle a,b | aBA^2,Ba^2bA \rangle\\ & \langle a,b | aBA^2,Ba^2bA \rangle \xrightarrow[Ba^2bA *= aBA^2]{} \langle a,b | B,aBA^2 \rangle 
\end{align*}

 \newpage
 \setcounter{equation}{0}
\textbf{T13}:
\begin{align*} 
& \langle a,b | a^2bAbAB,b^2aBaBA \rangle \xrightarrow[(b^2aBaBA)^A]{} \langle a,b | a^2bAbAB,ab^2aBaBA^2 \rangle\\ 
& \langle a,b | a^2bAbAB,ab^2aBaBA^2 \rangle \xrightarrow[ab^2aBaBA^2 *= a^2bAbAB]{} \langle a,b | ab,a^2bAbAB \rangle\\ & \langle a,b | ab,a^2bAbAB \rangle \xrightarrow[(ab)^A]{} \langle a,b | a^2bA,a^2bAbAB \rangle\\ 
& \langle a,b | a^2bA,a^2bAbAB \rangle \xrightarrow[(a^2bA)^{-1}]{} \langle a,b | aBA^2,a^2bAbAB \rangle\\ & \langle a,b | aBA^2,a^2bAbAB \rangle \xrightarrow[(aBA^2)^b]{} \langle a,b | BaBA^2b,a^2bAbAB \rangle\\ 
& \langle a,b | BaBA^2b,a^2bAbAB \rangle \xrightarrow[(a^2bAbAB)^b]{} \langle a,b | BaBA^2b,Ba^2bAbA \rangle\\ 
& \langle a,b | BaBA^2b,Ba^2bAbA \rangle \xrightarrow[BaBA^2b *= Ba^2bAbA]{} \langle a,b | A,Ba^2bAbA \rangle\\ 
%\textbf{shortened}
% & \langle a,b | A,Ba^2bAbA \rangle \xrightarrow[(A)^{-1}]{} \langle a,b | a,Ba^2bAbA \rangle\\ & \langle a,b | a,Ba^2bAbA \rangle \xrightarrow[(Ba^2bAbA)^b]{} \langle a,b | a,B^2a^2bAbAb \rangle\\ 
% & \langle a,b | a,B^2a^2bAbAb \rangle \xrightarrow[(a)^{-1}]{} \langle a,b | A,B^2a^2bAbAb \rangle 
\end{align*} 
% + ACMoveSeq(c(1,-1),m(1,0),c(0,-1),i(0),c(0,2),c(1,2),m(0,1),i(0),c(1,2),i(0))

%%%%%%%%%%%%%%%%%%%%%%%%%%%%%%%%%%%

\iffalse

%%%%%%%%%%%%%%%%%%%%%%%%%%%%%%%%%%%
% Original unshortened version of T13:

\textbf{T13}:\\
\begin{align*} & \langle a,b | a^2bAbAB,b^2aBaBA \rangle \xrightarrow[(b^2aBaBA)^A]{} \langle a,b | a^2bAbAB,ab^2aBaBA^2 \rangle\\ & \langle a,b | a^2bAbAB,ab^2aBaBA^2 \rangle \xrightarrow[ab^2aBaBA^2 *= a^2bAbAB]{} \langle a,b | ab,a^2bAbAB \rangle\\ & \langle a,b | ab,a^2bAbAB \rangle \xrightarrow[(ab)^A]{} \langle a,b | a^2bA,a^2bAbAB \rangle\\ & \langle a,b | a^2bA,a^2bAbAB \rangle \xrightarrow[(a^2bA)^{-1}]{} \langle a,b | aBA^2,a^2bAbAB \rangle\\ & \langle a,b | aBA^2,a^2bAbAB \rangle \xrightarrow[(aBA^2)^b]{} \langle a,b | BaBA^2b,a^2bAbAB \rangle\\ & \langle a,b | BaBA^2b,a^2bAbAB \rangle \xrightarrow[(a^2bAbAB)^b]{} \langle a,b | BaBA^2b,Ba^2bAbA \rangle\\ & \langle a,b | BaBA^2b,Ba^2bAbA \rangle \xrightarrow[BaBA^2b *= Ba^2bAbA]{} \langle a,b | A,Ba^2bAbA \rangle\\ & \langle a,b | A,Ba^2bAbA \rangle \xrightarrow[(A)^{-1}]{} \langle a,b | a,Ba^2bAbA \rangle\\ & \langle a,b | a,Ba^2bAbA \rangle \xrightarrow[(Ba^2bAbA)^b]{} \langle a,b | a,B^2a^2bAbAb \rangle\\ & \langle a,b | a,B^2a^2bAbAb \rangle \xrightarrow[(a)^{-1}]{} \langle a,b | A,B^2a^2bAbAb \rangle 
\end{align*}
%\end{figure}

%%%%%%%%%%%%%%%%%%%%%%%%%%%%%%%%%%%

\fi

%%%%%%%%%%%%%%%%%%%%%%%%%%%%%%%%%%%

\protect\caption{The sequences of trivializing moves found for T1 and T13. \label{fig:trivExamples}}

\end{figure*}
%JS: now padded this out with some speculation --- please sanity check
%KK: DONE. I like it, fixed references and split into two paragraphs
It is interesting to note that none of the rank\,3 presentations and
none of the `O' instances (i.e.\ those of unknown triviality) were
solved by this approach. Examining elementary differences between presentations does not provide any very helpful guidance: for example, the Hamming distance $H$ between (the first relators of) successfully
solved presentations T81 and T82 is 4, whereas the distance between T81
and the unsolved T83 is only 2. If one takes generators as being
equivalent to their inverses, then both $H(T81,T82)$ and $H(T81,T83)$
are zero. 

As observed by Havas and Ramsay \citep{HavasRamsay:2003}, relator length behaves highly nonmonotonically along the path to a solution. In general, the highly discontinuous effect of free reduction on words means that it is difficult to discern any distinguishing characteristics of the successful trivialization sequences. One might speculate that one of the main reasons that the ACC remains unsolved is that, considered in terms of algorithmic information theory \citep{CPLX:CPLX6130010410}, AC-trivializations are `nearly incompressible', i.e.\ cannot be readily expressed by a function of significantly lower complexity than the sequence itself. Pending deeper algebraic insights, this apparent lack of `obviously exploitable' structure lends further support for the learned bias of our approach.

\section{Conclusion}

The Andrews-Curtis conjecture is a longstanding open problem of interest
to topologists and group theorists \citep{AndrewsCurtis:1965}. Attempts
to eliminate potential counterexamples to the conjecture via combinatorial
search has seen no practical improvement since the exhaustive enumerative
approach of \citep{BowmanMcCaul:2006} in 2006. Informed by previous
work that analysed fitness correlations in the associated fitness
landscape \citep{doi:10.1142/S0218196711006753}, we generate new
predictors of search progress by performing offline learning to obtain good
fitness functions. These predictors take the form of random
walks in the search space that are good correlates for a more na\"{i}ve
measure of solution quality (i.e.\ total length of relators). This
is supplemented with an online approach that randomly samples a subset
of predictors. By this means, we successfully solved 19 problem instances which have withstood human and machine approaches since 2001.

Many solutions obtained by this approach comprise 20 or more moves and are thus substantially longer than the ones systematically enumerated in \citep{BowmanMcCaul:2006} (up to length 17). Assuming the effective branching factor of 8 (Section \ref{sec:Previous-Work}), a sequence of length 20 corresponds to a search tree of $8^{21}-1\approx 9.22\times 10^{18}$ nodes, arguably much too large to be systematically searched using algorithms like breadth-first search with currently available computational resources. For problem T56, with the longest trivializing sequence found in this study (25 moves), the tree is still greater by five orders of magnitude ($3\times 10^{23}$ nodes). Given the absence of universal fitness functions to efficiently guide the search \citep{doi:10.1142/S0218196711006753} reliance on some form of machine learning appears essential in obtaining further solutions combinatorially. %JS: %JS: I've reworded slightly; KK: Great, thx
%even if these numbers can be further reduced by considering more sophisicated properties of AC moves, 

%We expect that studies like this one and alike (e.g., \citep{Spector:2008:gecco} cited in Section \ref{sec:Previous-Work}) may rise the interest in metaheuristics and machine learning as sources of viable approaches to challenging mathematical problems. 
%As the sophistication of machine learning algorithms and available computational resources continue to grow, we believe that pure mathematics will provide an increasingly fruitful application area. 

% JS: TODO. I've added Spector but the last sentence is arguably a bit grandiose - could maybe lose it and/or find a better place to put the Spector reference.
% KK: I've tried to move that the reference to Spector to the end of Related Work. It is also not a perfect fit there, but IMO a bit better.  Also, tried to add a sentence before your 'grandiose' (and great BTW) sentence, but I'm not attached to this. If you find this last para of the paper forced, dont hesitate to remove it. 
% JS: I've killed it: let's save it for the text of the GECCO Humies submission...

\section*{Acknowledgments}

K. Krawiec acknowledges support from the National Science Centre (Narodowe Centrum Nauki) grant number 2014/15/B/ST6/05205. 

\bibliographystyle{abbrvnat}
\bibliography{andrews-curtis-distance-learning}

\end{document}